\documentclass[10pt,twocolumn,letterpaper]{article}

\usepackage[pagenumbers]{cvpr} 

\usepackage{algorithm}
\usepackage{algorithmic}
\usepackage{amsmath}
\usepackage{bm}
\usepackage{amssymb}
\usepackage{amsfonts}
\usepackage{cite} %
\usepackage{csquotes} %
\usepackage{physics}
\usepackage{booktabs}
\usepackage{multirow}
\usepackage{xcolor} 
\usepackage{threeparttable}
\usepackage{arydshln}
\usepackage{threeparttable}

\usepackage{makecell}
\usepackage{etoolbox}

\definecolor{cvprblue}{rgb}{0.21,0.49,0.74}
\usepackage[pagebackref,breaklinks,colorlinks,allcolors=cvprblue]{hyperref}

\def\paperID{8049} 
\def\confName{CVPR}
\def\confYear{2025}

\title{VQ4ALL: Efficient Neural Network Representation via a Universal Codebook}

\author{
   Juncan Deng$^{1,2}$~\thanks{Work done during an internship at vivo Mobile Communication Co., Ltd.} \quad Shuaiting Li$^{1,2}$\footnote[1]{}\quad Zeyu Wang $^{1}$ \\
   Hong Gu$^{2}$ \quad Kedong Xu$^{2}$  \quad Kejie Huang$^{1}$\\ 
   $^{1}$Zhejiang University  \quad \quad
   $^{2}$vivo Mobile Communication Co., Ltd \quad \quad \\
  \small\texttt{\{dengjuncan, list, wangzeyu2020, huangkejie\}@zju.edu.cn} \\
  \small\texttt{\{guhong, xukedong\}@vivo.com} \\
}

\begin{document}
\maketitle

\begin{abstract}

The rapid growth of the big neural network models puts forward new requirements for lightweight network representation methods. The traditional methods based on model compression have achieved great success, especially VQ technology which realizes the high compression ratio of models by sharing code words. However, because each layer of the network needs to build a code table, the traditional top-down compression technology lacks attention to the underlying commonalities, resulting in limited compression rate and frequent memory access. In this paper, we propose a bottom-up method to share the universal codebook among multiple neural networks, which not only effectively reduces the number of codebooks but also further reduces the memory access and chip area by storing static code tables in the built-in ROM. Specifically, we introduce VQ4ALL, a VQ-based method that utilizes codewords to enable the construction of various neural networks and achieve efficient representations. The core idea of our method is to adopt a kernel density estimation approach to extract a universal codebook and then progressively construct different low-bit networks by updating differentiable assignments. Experimental results demonstrate that VQ4ALL achieves compression rates exceeding 16 $\times$ while preserving high accuracy across multiple network architectures, highlighting its effectiveness and versatility.
\end{abstract}

\section{Introduction}
\label{sec:intro}
Neural networks have become indispensable tools in modern computer vision tasks due to their exceptional performance. However, these models are often resource-intensive, requiring substantial memory and computational power. Deploying them on resource-constrained platforms, where multiple networks may need to run simultaneously, remains a significant challenge. Addressing this limitation requires two critical advancements: enabling more universal network representation and more efficient network compression.



Traditional methods for reducing model size and enabling hardware acceleration typically employ techniques such as pruning and quantization. While these methods can achieve acceleration to some extent, they often suffer from significant performance degradation when aiming for extreme model compression. This is primarily because low-bit quantization introduces large quantization errors, often requiring mixed precision to maintain performance. Meanwhile, pruning leads to the loss of important weights, causing substantial structural differences in the pruned model. Both methods result in low hardware compatibility, making it challenging to design a unified accelerator architecture.


Vector quantization is a more hardware-friendly method of model compression that can achieve a higher compression ratio. However, traditional vector quantization requires training a separate codebook for each neural network, leading to long training times. Additionally, when achieving higher compression ratios, larger codebooks are often needed, resulting in longer loading times when switching between tasks. To address these challenges, we propose that it is essential to explore the underlying commonalities across neural networks. By leveraging these shared logical patterns, we can construct multiple neural network models in a bottom-up fashion. This approach not only effectively reduces the model size but also facilitates the design of unified low-level hardware, enabling general-purpose acceleration capabilities. Furthermore, it significantly reduces model compression time and holds the potential for enhancing the interpretability of neural networks.



In this paper, we introduce VQ4ALL, a VQ-based method that enables universal network representation and efficient network compression. VQ4ALL adopts a kernel density estimation approach to extract a universal codebook and then learn the differentiable assignments to gradually construct the low-bit neural network. The universal codebook can be stored in ROM, which significantly reduces the silicon area and eliminates the need for repeated codebook loading during fast task switching. Experimental results demonstrate that our approach not only achieves accuracy comparable to the original models across various networks but also significantly reduces storage requirements. Our main contributions can be summarized as follows:

\begin{itemize}
    \item We propose VQ4ALL, a novel VQ-based approach that enables universal network representation and efficient network compression.
    \item Our method introduces a kernel density estimation method to generate a universal codebook and a differentiable assignment learning strategy to progressively build low-bit neural networks.
    \item Our method achieves high compression rates (exceeding 16 $\times$) while maintaining accuracy comparable to original models across multiple architectures, demonstrating its effectiveness and versatility.
\end{itemize}


\begin{figure*}[!tbp]
  \centering
  \includegraphics[width=1\linewidth]{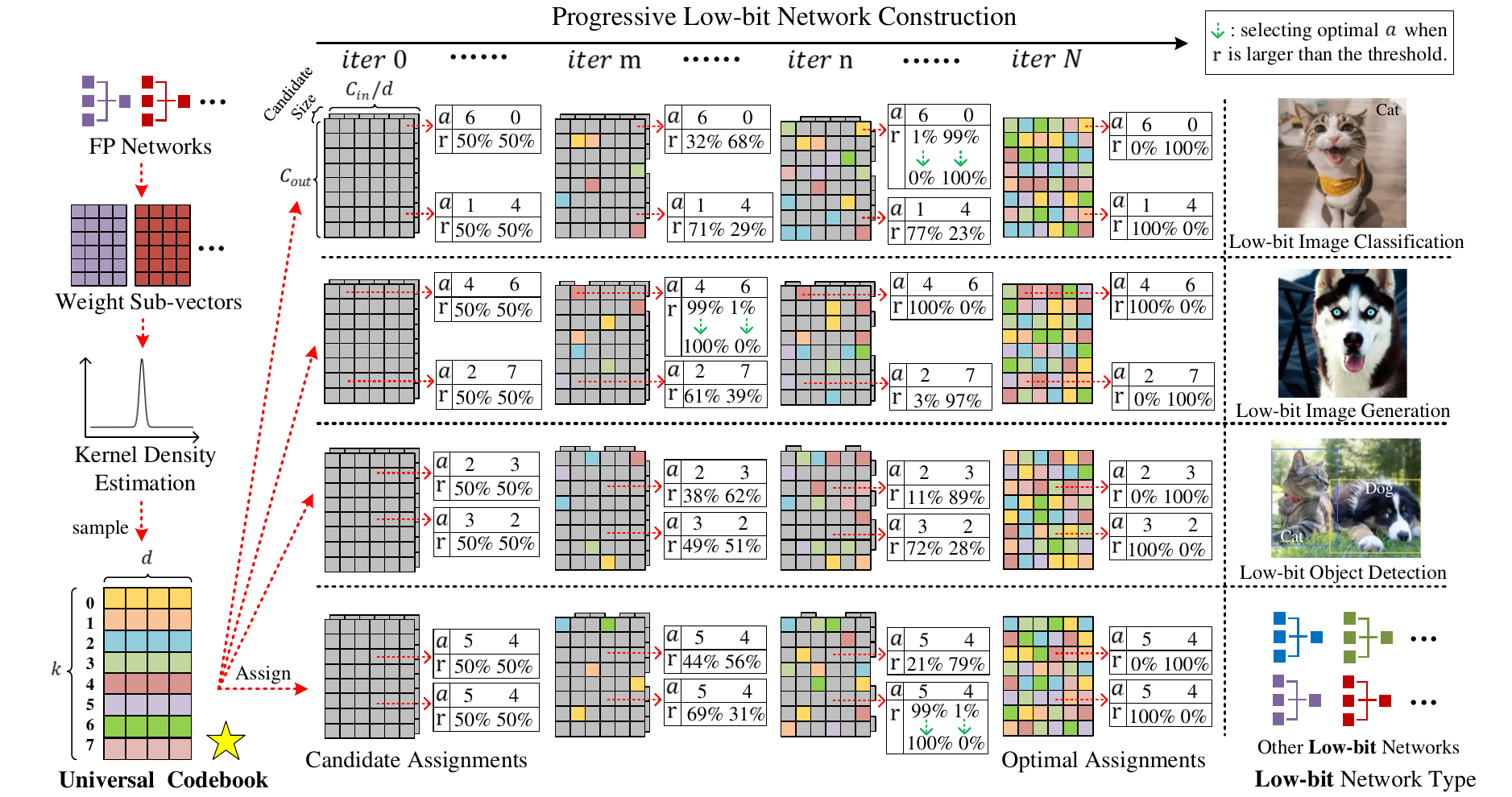}
    \caption{\textbf{The bottom-up pipeline of VQ4ALL.} The Universal Codebook is randomly sampled from the kernel density estimation of the floating-point sub-vectors of several networks. Based on the codebook, differentiable candidate assignments are assigned to each type of network, all initialized with the same ratios. The progressive network construction strategy then calibrates the ratios and sets candidate assignments with high ratios as optimal. Finally, various low-bit networks are constructed with the same universal codebook.}
  \label{pipeline}
\end{figure*}

\section{Related Work}
\label{sec:related}

Quantization is a commonly used method in network compression, which is mainly divided into two types: uniform quantization and vector quantization.



\textbf{Uniform Quantization.} Uniform quantization converts floating-point weights into integer values using scale coefficients, applied either tensor-wise, channel-wise, or group-wise. Most current methods focus on 8-bit or lower precision, often leveraging second-order information or quantization errors to enhance the process. To improve the accuracy of quantized models, training-time quantization techniques have been proposed. For instance, EWGS~\citep{lee2021network} adjusts gradients by scaling them based on Hessian approximations for each layer, while PROFIT~\citep{park2020profit} employs an iterative process and freezes layers depending on activation instability. Quant-Noise~\citep{quant_noise}, a structured dropout approach, applies quantization to a random subset of weights, allowing any quantization technique to be used. This approach enhances the predictive power of the compressed model. However, uniform quantization incurs a larger error at extremely low bit-width quantization due to its limitation of reconstructing weights in equidistant distributions, and the quantization coefficients cannot be shared across different DNNs.

\textbf{Vector Quantization.} Vector quantization (VQ) in neural networks was first introduced by Gong~\etal~\citep{gong2014compressing}, who explore scalar, vector, and product quantization (PQ) for fully connected (FC) layers. Wu~\etal~\citep{wu2016quantized} extend PQ to compress both FC and convolutional layers of CNNs by sequentially quantizing layers to reduce error accumulation. DeepCompression~\citep{deepcomp_iclr16} introduces k-means clustering for model compression by clustering weights and assigning the same value to all weights within a cluster. Son~\etal~\citep{son2018clustering} apply vector quantization to 3$\times$3 convolutions and fine-tune centroids via gradient descent, using extra bits to encode filter rotations, thus creating compact codebooks. HAQ~\citep{haq} utilizes reinforcement learning to search for the optimal quantization policy across various tasks, combining k-means clustering, similar to DeepCompression, with flexible bit-widths for different layers. BGD~\citep{stock2019and} uses PQ to compress both convolutional and FC layers through a clustering approach aimed at minimizing the reconstruction error of layer outputs, followed by end-to-end training of the centroids via distillation. PQF~\citep{martinez2021permute} applies rate-distortion theory to reorder weights, reducing clustering errors, and further compresses the network using an annealing algorithm. GOBO~\citep{gobo} first identifies outlier weights that are far from the average of each layer's weights and stores them uncompressed, while clustering the remaining weights using an algorithm similar to K-means. DKM~\citep{cho2021dkm} employs a differentiable K-means formulation integrated with the original loss function to recover the network's accuracy. VQ4DiT~\citep{deng2024vq4dit} suggests that fixed initial assignments lead to suboptimal updates of the codebooks, and therefore proposes a search algorithm for the most suitable assignments. Our method is an improvement and extension of VQ4DiT. Existing VQ-based methods primarily focus on generating per-layer codebooks in a top-down manner, particularly aiming to reduce clustering errors or preserve important weight information. However, this significantly increases the memory usage and access frequency of the codebooks and prevents the sharing of codebooks across different networks.


\section{Motivation}

\subsection{Types of quantization}
We provide a concise overview of uniform quantization and vector quantization.

\textbf{Uniform Quantization:} Let $ W \in \mathbb{R}^{o \times i} $ represent the weight matrix, where $o$ denotes the number of output channels and $i$ is the number of input channels. A standard symmetric uniform quantizer approximates the original floating-point weight $W$ as $ \widehat{W} \approx s W_{int} $, where $W_{int}$ is a $b$-bit integer representation and $s$ is a high-precision scaling factor shared across all elements of $W$. However, uniform quantization incurs a larger error at very low bit widths due to its inherent limitation in reconstructing weights with equidistant distributions.

\textbf{Vector Quantization:} A more flexible quantization approach is vector quantization (VQ), which expresses $W$ in terms of assignments $A$ and a codebook $C$.
$W$ is divided into $d$ dimension sub-vectors $w_{i,j} \in \mathbb{R}^{1 \times d}$:
\begin{equation}
  W =
  \begin{bmatrix}
      w_{1,1} & w_{1,2} & \cdots & w_{1,i/d} \\
      w_{2,1} & w_{2,2} & \cdots & w_{2,i/d} \\
      \vdots & \vdots & \ddots & \vdots \\
      w_{o,1} & w_{o,2} & \cdots & w_{o,i/d}
  \end{bmatrix},
\end{equation}
where $o \cdot i/d$ is the total number of sub-vectors. 
These sub-vectors are represented by a codebook
$C = \{ c(1), \ldots, c(k) \} \subseteq \mathbb{R}^{d \times 1}$, where $c(k)$ is referred to as the $k$-th codeword of length $d$.
The assignments $A = \{a_{i,j}\in \{1, \ldots, k\}\}  $ are the indices of each codeword that best represent every sub-vectors $\{ w_{o,i/d} \}$. 
The quantized weight $ \widehat{W} $ is reconstructed by replacing each $ w_{i,j} $ with $ c(a_{i,j}) $:
\begin{equation}
  \widehat{W} = C[A] =
  \begin{bmatrix}
      c(a_{1,1}) & c(a_{1,2}) & \cdots & c(a_{1,i/d}) \\
      c(a_{2,1}) & c(a_{2,2}) & \cdots & c(a_{2,i/d}) \\
      \vdots & \vdots & \ddots & \vdots \\
      c(a_{o,1}) & c(a_{o,2}) & \cdots & c(a_{o,i/d})
  \end{bmatrix}.
\label{cfunction}
\end{equation}
$A$ can be stored using $ \frac{o \times i}{d} \times \log_2 k $ bits and $C$ can be stored using $k \times d \times 32$ bits.

\begin{table}[tbp]
\small
\centering
\begin{tabular}{ccccccc}
    \toprule
    Bit  & $k,d$ & Type & $C$ & MSE & Rate  & I/O\\
    \midrule
    \multirow{3}{*}{\shortstack{3}}  
      & -- & UQ & -- & 2.8e-4  & -- & -- \\
      \cmidrule(lr){2-7}
      & $2^6,2$ & P-VQ & 264K & 5.3e-5  & 10$\times$ & 514$\times$ \\
      & $2^{12},4$ & U-VQ & \textbf{64K} & 1.8e-4  & \textbf{11$\times$} & \textbf{1$\times$} \\
    \midrule
    \multirow{3}{*}{\shortstack{2}} 
      & -- & UQ & -- & 1.1e-3  &-- & -- \\
      \cmidrule(lr){2-7}
      & $2^8,4$ & P-VQ & 2.1M & 1.2e-4  & 15$\times$ & 514$\times$ \\
      & $2^{16},8$ & U-VQ & \textbf{2.0M} & 2.1e-4  & \textbf{16$\times$} & \textbf{1$\times$} \\
    \midrule
    \multirow{3}{*}{\shortstack{1}}  
      & -- & UQ & -- & 4.8e-3  & -- & -- \\
      \cmidrule(lr){2-7}
      & $2^8,8$ & P-VQ & 4.2M & 3.2e-4  & 27$\times$ & 514$\times$ \\
      & $2^{16},16$ & U-VQ & \textbf{4.0M} & 4.4e-4  & \textbf{32$\times$} & \textbf{1$\times$} \\
    \bottomrule
\end{tabular}
\caption{\textbf{Metrics of classic uniform quantization (UQ) and vector quantization (VQ)} across a range of networks, including ResNet-18/50, Stable Diffusion V1-4, MobileNet-V1/2, and Mask-RCNN R-50 FPN. 'Bit' indicates the ideal quantization bit width. The dimensions of the codebook for VQ are represented as $k \times d$. VQ is categorized into two types: 'P-VQ', where per-layer is constructed using an independent codebook, and 'U-VQ', where all networks are constructed using a universal codebook. $C$ denotes the memory usage of all codebooks. 'MSE' denotes the mean square quantization error. 'I/O' refers to the number of times the codebooks are accessed.}
\label{qerror}
\end{table}

\subsection{Benefits of a universal codebook}
As illustrated in Table~\ref{qerror}, we apply the classic uniform quantization (UQ) and vector quantization (VQ) to the multiple networks. At the same bit-width, per-layer VQ (P-VQ) results in a much smaller quantization error compared to UQ. Increasing $k$ and $d$ while maintaining constant memory usage for assignments effectively reduces quantization error. However, larger values of $k$ and $d$ result in longer preprocessing time and extended calibration iterations, further complicating the process. Moreover, when per-layer VQ is simultaneously applied to multiple networks on a computing platform, it significantly increases the memory requirements and I/O times of codebooks, which poses a substantial challenge.

Our method introduces a new VQ type, called universal VQ (U-VQ), where all networks share a single universal codebook. The storage cost of the codebook is evenly distributed, enabling extremely high compression rates. At the same time, the codebook is fitted from the weights of multiple networks, with negligible preprocessing time. The quantization noise of Universal VQ remains significantly smaller than that of Uniform Quantization (UQ) and is on par with Per-layer VQ. Furthermore, the universal codebook can be stored in ROM, reducing silicon area and eliminating the need for repeated codebook loading during rapid task switching. Constructing different neural networks based on a universal codebook opens the possibility of designing a unified accelerator architecture, thereby reducing hardware overhead caused by data transformations.

\section{VQ4ALL}
To address the identified challenges, we introduce VQ4ALL, a VQ-based method that utilizes a universal codebook to enable the construction of various low-bit neural networks and achieve efficient representations. The pipeline of VQ4ALL is visualized in Figure \ref{pipeline}. In Section 4.1, we describe the initialization of the universal codebook, which applies to most deep neural networks. In Section 4.2, we introduce the definition of the objective function. In Section 4.3, we explain how the network is progressively constructed through the learning of differentiable candidate assignments.

\subsection{Initialization of the Universal Codebook}
To generate a universal shared codebook, the normal approach is to concatenate the weight sub-vectors of multiple models and apply the K-means clustering method to obtain the codebook. However, this approach is time-consuming and the resulting codebook may only be effective for a few models with a large number of parameters. 
Therefore, we randomly sample an equal number of weight sub-vectors from each network and concatenate them, ensuring that the codebook remains unbiased. We then use the kernel density estimation (KDE) algorithm to analyze the distribution of these sub-vectors:
\begin{equation}
    f(w) = \frac{1}{n h} \sum_{i=1}^{n} K\left(\frac{w - w_{o,i/d}}{h}\right), K(u) = \frac{1}{\sqrt{2\pi}} e^{-\frac{u^2}{2}},
\end{equation}
where $f(w)$ represents the estimated density function, $n$ is the number of sub-vectors, $w_{o,i/d}$ are the sub-vectors being analyzed, $h$ is the bandwidth that controls the smoothing of the estimation, and $K(u)$ is the Gaussian kernel function. The computation process of KDE is highly efficient, and $\hat{f}(w)$ can be used to randomly sample a frozen universal codebook of size $k \times d$ :
\begin{equation}
    C = \{ c(1),\ldots, c(k)\}, c(k) \sim f(w).
\end{equation}
The next step is to determine which codeword to use for representing each weight sub-vector. The most straightforward approach is to select the codeword with the smallest distance. However, this method can introduce significant quantization errors, which may result in a considerable loss of network accuracy.
For each weight sub-vector, we calculate its Euclidean distance to all codewords, obtaining the indices of the top $n$ closest codewords:
\begin{equation}
  A_c = \{a_{o,i/d}\}_n = \arg\min_{k}^n \| w_{o,i/d} - c(k) \|_2^2,
\end{equation}
where $A_c$ is the candidate assignments of the sub-vectors and $n$ is the number of each $A_c$. We need to select the most appropriate assignments from the candidate assignments to effectively represent the information. To achieve this, we assign softmax ratios $R$ to all candidate assignments:
\begin{equation}
  R = \{r_{o,i/d}\}_n = \{\frac{e^{\{{z_{o,i/d}\}_n}}}{\sum_{j=1}^n e^{\{{z_{o,i/d}\}_n}}}\}_n, \sum_n \{r_{o,i/d}\}_n = 1,
\end{equation}
where $\{z_{o,i/d}\}_n$ is the actual value of each ratio, which is initialized to be inversely proportional to the Euclidean distance:
\begin{equation}
    \{z_{o,i/d}\}_{m \in n} = \ln{ \frac{\| w_{o,i/d} - c(\{a_{o,i/d}\}_{-1}) \|_2^2}{\| w_{o,i/d} - c(\{a_{o,i/d}\}_m) \|_2^2} }.
\end{equation}
Therefore, $\widehat{W} = RC[A_c]$ can be reconstructed based on the differentiable weighted average function:
\begin{equation}
    \begin{bmatrix}
        \{r_{1,1}\}_n \times c(\{a_{1,1}\}_n)  & \cdots & \{r_{1,i/d}\}_n \times c(\{a_{1,i/d}\}_n) \\
        \{r_{2,1}\}_n \times c(\{a_{2,1}\}_n)  & \cdots & \{r_{2,i/d}\}_n \times c(\{a_{2,i/d}\}_n) \\
        \vdots  & \ddots & \vdots \\
        \{r_{o,1}\}_n \times c(\{a_{o,1}\}_n) &  \cdots & \{r_{o,i/d}\}_n \times c(\{a_{o,i/d}\}_n)
    \end{bmatrix}
\end{equation}

\subsection{Definition of Objective Function}
Our method aims to construct multiple low-bit deep neural networks that exhibit the same behavior as the floating-point networks. Notably, the universal codebook remains unchanged throughout. We combine three common training strategies to effectively update candidate assignments and other network parameters (e.g., bias and normalization layers). Specifically, given the input $\mathbf{x}$ and the target $\mathbf{y}$ from the calibration dataset for the quantized network $\mathbf{\epsilon}_q$, the task objective function is computed as:
\begin{equation}
    \mathcal{L}_t = \mathbb{E}_{\mathbf{x},\mathbf{y},\mathbf{\epsilon}_q} \left[ \left\| \mathbf{y} - \mathbf{\epsilon}_q(\mathbf{x}) \right\|_2^2 \right].
\end{equation}

To ensure that the output of each network block in $\mathbf{\epsilon}_q$ is similar to that of the floating-point network $\mathbf{\epsilon}_{fp}$, we also apply block-wise knowledge distillation (KD) as a constraint. Given the same input $\mathbf{x}$ to both $\mathbf{\epsilon}_{fp}$ and $\mathbf{\epsilon}_{q}$, the KD objective function is computed as:
\begin{equation}
    \mathcal{L}_{kd} = \mathbb{E}_{\mathbf{x},\mathbf{b}_{fp},\mathbf{b}_q} \left[ \sum_{l} \left\| b_{fp}^l(\mathbf{x}) - b_q^l(\mathbf{x}) \right\|_2^2 \right],
\end{equation}
where $b_{fp}^l$ and $b_q^l$ represent the features of $l$-th main block from $\mathbf{\epsilon}_{fp}$ and $\mathbf{\epsilon}_q$, respectively. 

To accelerate the update of $R$ towards values close to 0 or 1, rather than getting stuck in local optima, we introduce a regularization function for $R$:
\begin{equation}
    \mathcal{L}_r =  n \times \sum_{o,i/d,n} ( \{r_{o,i/d}\}_n \times (1 - \{r_{o,i/d}\}_n) / ( \frac{o \times i}{d} ).
\end{equation}

The final objective function $\mathcal{L}$ is represented as: 
\begin{equation}
    \mathcal{L} = \mathcal{L}_t + \mathcal{L}_{kd} + \mathcal{L}_{r}.
\end{equation}
During the above process, we update the ratios through gradients: $R \leftarrow R - u \left( \pdv{\mathcal{L}}, \theta \right)$, where $u$ is the optimizer with hyperparameters $\theta$.

\subsection{Progressive Network Construction Strategy}
As shown in Figure \ref{pnc} (Down), most of the largest ratios oscillate around 1, while a small number of them fluctuate within certain ranges in the later stages of the updating process. Therefore, selecting the optimal assignments to construct low-bit networks becomes challenging. DKM addresses this issue by enforcing a transition from soft assignments to hard assignments. We replicate this approach by selecting the candidate assignments with the highest ratios as the optimal assignments. However, as shown in Figure \ref{pnc} (Up), this approach significantly reduces the accuracy of the low-bit network due to discrepancies introduced by the reconstructed weights:
\begin{equation}
    \sum \left\| RC[A_c] - C[A_c[R_{max}]] \right\|_2^2 > 0.
\end{equation}
Such discrepancies cause corresponding deviations in the features of each layer of the deep neural network, which eventually accumulate in the output.

Therefore, we propose the Progressive Network Construction (PNC) strategy, a gradual approach that prevents network collapse. When updating the parameters, once the ratio of a candidate assignment becomes very close to 1, PNC adopts a frozen one-hot mask to set it as the optimal assignment with its ratio fixed at 1 and other ratios fixed at 0. The formula can be expressed as:
\begin{equation}
     \{r_{o,i/d}\}_n = \text{one-hot}(m), \{r_{o,i/d}\}_{m \in n}>\alpha,
\end{equation}
where $\alpha$ is a specific value (e.g. 0.9999). Additionally, $\mathcal{L}_{r}$ is computed only for the unset ratios. Compared to selecting all optimal assignments at once, the accuracy loss from the PNC is much smaller, allowing the remaining candidate assignments larger space to restore the model's accuracy. In this way, the network is gradually constructed.

\begin{figure*}[t]
  \centering
  \includegraphics[width=1\linewidth]{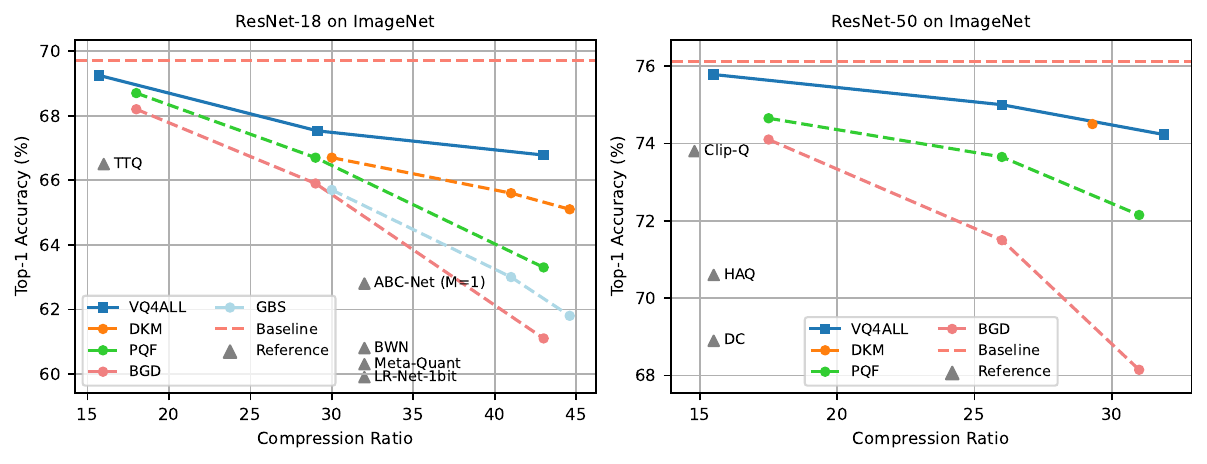}
    \caption{\textbf{Compression results for ResNet-18 and ResNet-50.} We compare the trade-off between accuracy and compression ratio, using pre-trained models from the PyTorch zoo as baselines. Overall, our method demonstrates superior accuracy compared to previous approaches.}
  \label{compare}
\end{figure*}

\begin{table*}[tbp]
\small
\centering
\begin{tabular}{lcc |ccc |ccc}
    \toprule
    & Size~& Ratio~&
    AP\textsuperscript{bb} & AP$_{50}^{\text{bb}}$ & AP$_{75}^{\text{bb}}$ & AP\textsuperscript{mk} & AP$^{\text{mk}}_{50}$  & AP$^{\text{mk}}_{75}$ \\
    \midrule
    Mask-RCNN R-50 FPN (uncompressed)   & 169.40 MB &           1.0$\times$ & 37.9 & 59.2 & 41.1 & 34.6 & 56.0 & 36.8 \\
    RetinaNet (uncompressed)          & 145.00 MB &           1.0$\times$ & 35.6 &   -- &   -- &   -- &   -- &   -- \\
    FQN                              &  18.13 MB &  8.0$\times$ & 32.5 & 51.5 & 34.7 &   -- &   -- &   -- \\ %
    HAWQ-V2                       &  17.90 MB &  8.1$\times$ & 34.8 &   -- &   -- &   -- &   -- &   -- \\
    \midrule
    BGD                              &   6.65 MB & 25.5$\times$ & 33.9 & 55.5 & 36.2 & 30.8 & 52.0 & 32.2 \\
    PQF        &   6.65 MB & 25.5$\times$  & 36.3 & 57.9 & 39.4 & 33.5 & 54.7 & 35.6 \\ 
    VQ4ALL             &   \textbf{6.44 MB} & \textbf{26.3$\times$}   & \textbf{36.8}  & \textbf{58.2} & \textbf{40.2} & \textbf{33.7} & \textbf{54.8} & \textbf{35.8} \\ 
    \bottomrule
\end{tabular}
\caption{
    \textbf{Object detection results on MS COCO 2017.}
    Results are reported for a compressed Mask R-CNN with a ResNet-50 backbone, along with other baseline object detection architectures. We present model size, compression ratio, and accuracy metrics, including bounding box (bb) and mask (mk) average precision (AP) at different IoU thresholds where available.
}
\label{maskrcnn}
\end{table*}

\section{Experiments}
We evaluate our method on ResNet-18/50~\citep{he2016deep} and MobileNet-V2~\citep{mobilenet_v2} for image classification, Mask R-CNN R-50 FPN~\citep{he2017mask} for object detection, and Stable Diffusion v1-4~\citep{rombach2022high} for image generation. The network weights and datasets are obtained through official sources. Our primary objective is to explore the trade-off between model compression and precision; therefore, activation quantization is disabled in all experiments.

\textbf{Hyperparameters:} All experiments are conducted on a single NVIDIA A6000 GPU with 48GB memory. The ratios of candidate assignments are fine-tuned by the Adamax optimizer with a learning rate of $3 \times 10^{-1}$ (without individual hyper-parameter tuning). The universal codebook sizes $k \times d$ used for 3/2/1/0.5 bit quantization are $2^{12} \times 4$, $2^{16} \times 8$, $2^{16} \times 16$, and $2^{16} \times 32$, respectively. To align with the configurations of other compression methods, the per-layer codebook sizes used for 2/1 bit quantization of special layer (e.g., the last output layer of image classification networks) are $2^{8} \times 4$, $2^{8} \times 8$, respectively. The number of candidate assignments in all cases of our method is set to 64.

\textbf{Universal Codebook:} We randomly extract $10 \times k \times d$ weight sub-vectors from each of the networks mentioned above and concatenate them. A kernel density estimation with a bandwidth of 0.01 is then applied efficiently to these concatenated sub-vectors. The resulting estimation is used to sample a single frozen universal codebook, which is used to construct most layers across different networks in our method.

\begin{table}[tbp]
\small
\centering
\begin{tabular}{cc|cccccc}
\toprule
   \multicolumn{2}{c}{}   & ResNet-18  & ResNet-50  & MobileNet  \\  
   \midrule
    Bit &    Base           & 69.8 / --     & 76.1 / --      &  71.9 / --   \\  
    \midrule
    \multirow{3}{*}{{\rotatebox[origin=c]{90}{3 bit}}} 
    &EWGS     & \textbf{70.5} / -- & 76.3  / --  & 64.5  / -- \\   
    &DKM       & 69.9  / 10$\times$  &  76.2  / 10$\times$       &  \textbf{70.3}  / 9$\times$ \\  
    &VQ4ALL                     & 69.8 / \textbf{11$\times$ }    & \textbf{76.3} / \textbf{11$\times$}      & 69.7 / \textbf{11$\times$} \\ 
    \midrule
    \multirow{3}{*}{{\rotatebox[origin=c]{90}{2 bit}}} 
    & EWGS     &   69.3 / -- & 75.8  / --  &  49.1  / -- \\  
    & DKM          &  68.9  / 15$\times$ &  75.3  / 15$\times$     &   66.2 / 14$\times$ \\  
    & VQ4ALL    & \textbf{69.4} / \textbf{16$\times$}    & \textbf{75.9}  / \textbf{16$\times$ }        &  \textbf{67.2} / \textbf{16$\times$}  \\  
    \midrule
    \multirow{3}{*}{{\rotatebox[origin=c]{90}{1 bit}}} 
    & EWGS      & 66.6 / -- &  73.8 / -- &  23.0 / --  \\ 
    & DKM        & 67.0 / 30$\times$ &  73.8 / 29$\times$      &   55.0 / 28$\times$     \\ 
    &VQ4ALL      & \textbf{68.0} / \textbf{32$\times$}    & \textbf{74.7}  / \textbf{32$\times$}       & \textbf{60.4}  / \textbf{32$\times$} \\  
    \bottomrule
   
\end{tabular} 
\caption{ \textbf{Image classification results on ImageNet.}
Under the same quantization configuration, we compare the Top-1 accuracy and compression ratio of our method with state-of-the-art quantization methods for image classification networks. Our approach maintains excellent accuracy and stability, even in extremely low-bit scenarios.}
\label{imagenetresult}
\end{table}

\subsection{Image Classification}
We benchmark our method on the task of image classification by compressing the popular ResNet-18/50 and MobileNet-V2 networks using the ImageNet dataset~\citep{deng2009imagenet}. We compare our VQ4ALL method with previous works: Trained Ternary Quantization (TTQ)~\citep{zhu2016trained}, Binary Weight Network (BWN)~\citep{rastegari2016xnor}, ABC-Net~\citep{lin2017towards}, LR-Net~\citep{shayer2017learning}, GBS~\citep{jang2016categorical}, DeepCompression (DC) ~\citep{deepcomp_iclr16}, Hardware-Aware Automated Quantization (HAQ)~\citep{haq}, CLIP-Q~\citep{frederick2018deep}, Meta-Quant~\citep{meta_quant}, "And The Bit Goes Down" (BGD)~\citep{stock2019and}, "Permute, Quantize, and Fine-tune" (PQF)~\citep{martinez2021permute}, and "Differentiable k-Means Clustering Layer" (DKM)~\citep{cho2021dkm}. All results presented are sourced from the original papers or additional surveys. We conduct extensive comparisons with the recently proposed DKM method, as it represents the state of the art on the compression of image classification tasks. While DKM shares some similarities with our approach, it does not address issues related to excessive memory usage, slow updates during training, and accuracy drops during inference. 

We set a batch size of 256 for all cases and run for 10 epochs. The learning rate for network parameters other than candidate assignments is set to 1e-3, using the Adam optimizer~\citep{kingma2014adam} and cosine annealing scheduler~\citep{loshchilov2016sgdr}. We do not compress the input layers (as they account for less than 0.05\% of the total network size), the biases, or the batch normalization layers. 
The output layer is constructed using a small per-layer codebook derived from clustering its weights, while other layers are constructed using the universal codebook.

As visualized in Figure \ref{compare}, our VQ4ALL significantly outperforms other methods across all compression ratios. For ResNet-18, VQ4ALL achieves a stable trade-off between compression ratio and accuracy: at a 16$\times$ compression ratio, VQ4ALL attains performance close to the floating-point network, surpassing TTQ by 3\% in Top-1 accuracy. At a compression ratio approaching 30$\times$, VQ4ALL still maintains an accuracy of 67.6\%, while DKM suffers accuracy loss due to forced assignment transitions, and PQF and BGD achieve suboptimal per-layer codebooks due to inconsistent gradients from sub-vectors. When the compression ratio exceeds 40$\times$, VQ4ALL continues to hold an accuracy close to 67\% because the memory footprint of the universal codebook is negligible, whereas the accuracy of other methods falls below 66\%. The results for ResNet-50 show a similar trend: even when the compression ratio exceeds 30$\times$, VQ4ALL achieves substantial gains across all compression ratios, particularly maintaining accuracy comparable to BGD and PQF at a 17.5$\times$ compression ratio.

We also compare our VQ4ALL algorithm with the state-of-the-art uniform quantization approach, EWGS~\citep{lee2021network}, which has outperformed previous approaches in low-precision settings. Table~\ref{imagenetresult} summarizes our comparison results on ResNet18, ResNet50, and MobileNet-v2 for the ImageNet classification task. Following the configuration of EWGS, none of the experiments in Table~\ref{imagenetresult} compress the input and output layers of the network. In 2-bit quantization, VQ4ALL outperforms all other methods, particularly achieving an 18.1\% accuracy improvement over EWGS on MobileNet-V2. With 1-bit quantization, VQ4ALL's advantage becomes even more pronounced, surpassing DKM in accuracy by nearly 1\% on ResNet-18/50. On the challenging-to-compress MobileNet-V2, VQ4ALL maintains an accuracy of 60.4\%. We also report the average compression rate of compressed layers. As a per-layer VQ-based method, DKM experiences increasing compression performance degradation in extremely low-bit quantization due to the memory overhead of codebooks. In contrast, VQ4ALL achieves optimal compression rates, as its universal codebook is shared with various low-bit networks and stored as static code tables in the built-in ROM.

\subsection{Object Detection and Segmentation}
We benchmark our method on the task of object detection and segmentation by compressing the popular Mask-RCNN R-50 FPN architecture using the MS COCO 2017 dataset~\citep{lin2014microsoft}. We also compare with other object detection baselines, including the Fully Quantized Network (FQN)~\citep{li2019fully} and the second version of Hessian Aware Quantization (HAWQ-V2)~\citep{dong2020hawq}, both of which demonstrate results on compressing RetinaNet~\citep{lin2017focal}. 

We use a batch size of 2 to update network parameters over 4 epochs. The optimizer and scheduler for other parameters are Adam and cosine annealing, respectively, with a learning rate of 5e-5. We follow the configurations of BGD and PQF by excluding certain layers from compression. 

As shown in Table \ref{maskrcnn}, our VQ4ALL achieves a box AP of 36.8\% and a mask AP of 33.7, surpassing the previous best VQ-based work, PQF, by 0.5\% and 0.2\%. Furthermore, VQ4ALL closely matches the performance of Mask-RCNN R-50 FPN and outperforms RetinaNet which has a size of 145MB, with a compressed size of only 6.44MB. Compared to other UQ-based and VQ-based methods, VQ4ALL maintains stability across all metrics even at extremely high compression ratios. We also attempted to replicate DKM by training a $2^8 \times 8$ per-layer codebook for each compressed layer but encountered out-of-memory issues. Therefore, we are unable to provide compression results for DKM on Mask-RCNN R-50 FPN. We found that DKM utilized 16 NVIDIA V100 GPUs for network compression with 200 epochs, which is impractical in most cases. In contrast, VQ4ALL only requires a single A6000 GPU to ensure generalizability.

\begin{table}[tbp]
\small
\centering
\begin{tabular}{cccccc}
\toprule
Method & Bit & FID$\downarrow$  & sFID$\downarrow$ & IS$\uparrow$ & CLIP*$\uparrow$ \\ 
\midrule
Base  & 32 & 25.44 & 85.34 & 41.32 & 26.46\\
\midrule
Q-diffusion & 3 & 65.13 & 94.38 & 14.46 & 18.71\\
PCR & 3 & 55.13 & 90.11 & 16.84 & 22.40\\
PQF & 3 & 28.40 & 88.95 & 31.98 & 24.73\\
VQ4ALL  & 3 & \textbf{26.40} & \textbf{87.42} & \textbf{37.02} & \textbf{25.96}\\

\midrule
Q-diffusion & 2 & 248.50 & 140.54 & 3.43 & 10.56\\
PCR & 2 & 238.83 & 134.43 & 3.77 & 12.71\\
PQF & 2 & 45.13 & 89.62 & 25.49 & 22.91\\
VQ4ALL  & 2 & \textbf{33.82} & \textbf{89.01} & \textbf{29.35} & \textbf{23.76}\\

\bottomrule

\end{tabular}
\caption{\textbf{Image generation results on COCO Prompts.} We evaluate the $512 \times 512$ generated images of Stable Diffusion v1-4 against the entire validation set. 'CLIP*' denotes CLIP score~\citep{hessel2021clipscore, radford2021learning}. We compute Fréchet Inception Distance (FID)~\citep{heusel2017gans}, spatial FID (sFID), and Inception Score (IS)~\citep{salimans2016improved} to assess visual quality. Additionally, we measure CLIP score with CLIP-ViT-g/14 model to assess text-image correspondence.}
\label{sd14_results}
\end{table}

\subsection{Image Generation}
We benchmark our method on the task of image generation by compressing the Stable Diffusion v1-4 using the MS COCO 2017 dataset~\citep{lin2014microsoft}. We compare our VQ4ALL with state-of-the-art quantization methods for diffusion models, including Q-diffuison~\citep{li2023q} and "Progressive Calibration and Activation Relaxing" (PCR)~\citep{tang2023post}.

We follow the standard training process for diffusion models, setting the batch size to 4 and the number of iterations to 10,000. We use the Adam optimizer with a learning rate of 1e-5 for other network parameters, without applying any learning rate updates. 

The image generation results are presented in Table \ref{sd14_results}. 
In 3-bit quantization, our VQ4ALL achieves a 28.73 reduction in FID and an 20.18 increase in IS compared to PCR, while maintaining a CLIP score similar to the baseline. As a VQ-based method, PQF outperforms Q-diffusion and PCR but still lags behind VQ4ALL. In 2-bit quantization, Q-diffusion and PCR fail to produce acceptable results due to significant quantization errors. VQ4ALL still achieves excellent performance, with the FID only 8.38 higher than the baseline, and the PQF 19.69 higher.

\begin{table}[tbp]
\small
\centering
\begin{tabular}{c|ccccccc}
\toprule
    $n$ & 1 & 8 & 64 (Our)  &  256  &  1024 \\ 
    \midrule
    Acc & 65.9 &    69.2       &  \textbf{69.4 }    & 69.4      &  oom   \\  
    \midrule
    \midrule
    Part & No $\mathcal{L}_t$ & No $\mathcal{L}_{kd}$ & No $\mathcal{L}_r$  & No PNC & Our  \\  
   \midrule
    Acc & 68.2 &    67.6       & nc      & 54.8      &  \textbf{69.4}   \\  
    \midrule
    \midrule
    Index & 0$\sim$11 &  12$\sim$23 &  24$\sim$35  &   36$\sim$47  &   47$\sim$63 \\ 
    \midrule
    \% & 83.1\% &    9.2\%       & 3.8\%      & 2.2\%      &  1.7\%   \\
    \bottomrule
   
\end{tabular} 
\caption{ \textbf{Ablation study on Image classification results of 2-bit ResNet-18.}
We perform ablations on the number of candidate assignments ($n$), the effect of each part of the VQ4ALL pipeline, and the index distribution of optimal assignments. ‘nc’ indicates non-convergence. 'oom' denotes that the compression process is out of memory.}
\label{ablation1}
\end{table}

\subsection{Ablation Study} 
In Table~\ref{ablation1}, we show an ablation study on image classification results of 2-bit ResNet-18. We first perform an ablation on the number of candidate assignments ($n$) for constructing optimal low-bit networks with minimal training time and memory usage. When $n=1$, VQ4ALL degenerates to a standard VQ-based method without codebook updates, achieving only 65.9\% accuracy. When $n=8$, accuracy sharply increases by 3.3\%, indicating that VQ4ALL effectively identifies optimal assignments and constructs optimal low-bit networks through different combinations of codewords within the same universal codebook. As $n$ continues to increase, the improvements in accuracy diminish, with $n=64$ nearly reaching the optimal performance of our VQ4ALL.

\begin{figure}[tbp]
  \centering
  \includegraphics[width=1\linewidth]{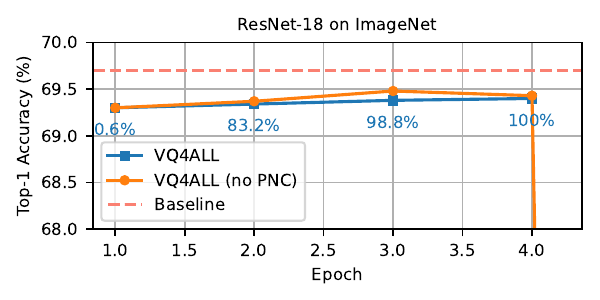} 
  \includegraphics[width=1\linewidth]{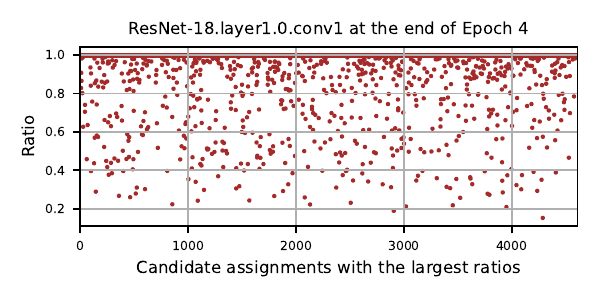} 
    \caption{\textbf{Ablation results on Progressive Network Construction (PNC) Strategy} with the same 2-bit ResNet-18 compression configuration. Up: we compare the accuracy of VQ4ALL with and without PNC at each epoch. Once the VQ4ALL pipeline is complete, we convert the candidate assignments with the largest ratios in VQ4ALL (no PNC) to optimal assignments, resulting in a drop in accuracy to 57.77\%. Down: in the distribution of the largest ratios, 15\% are outliers significantly distant from 1.}
  \label{pnc}
\end{figure}

We also evaluate the importance of each component in the VQ4ALL pipeline by measuring the accuracy after disabling each component. When block-wise distillation between the floating-point and low-bit networks is disabled, the accuracy of the low-bit network decreases by nearly 2\%, indicating that $\mathcal{L}_{kd}$ helps prevent significant deviations in the reconstructed weights and block outputs, thereby facilitating the learning of candidate assignments. Disabling $\mathcal{L}_{r}$ prevents the objective function from converging, demonstrating the necessity of regularizing $R$. The Progressive Network Construction (PNC) strategy is the most crucial component of the VQ4ALL pipeline. As shown in Figure \ref{pnc}, when PNC is disabled, VQ4ALL simultaneously converts all candidate assignments with the highest $R$ values to optimal assignments, leading to cumulative errors as described in Equation 13 and causing the accuracy to drop to 54.77\%. PNC ensures that VQ4ALL maintains the stability of network performance throughout the process of gradually constructing low-bit networks, thereby preventing performance deviation during network inference.

We further analyze the frequency distribution of each optimal assignment within the candidate assignments in Table~\ref{ablation1}. This distribution resembles a normal distribution, indicating that candidate assignments closer to the original sub-vectors are more likely to be optimal.

\section{Conclusion}
In this work, we propose a novel bottom-up approach for sharing a universal codebook across multiple neural networks. This method not only effectively reduces the number of required codebooks, but also minimizes memory access and chip area by storing static code tables in the built-in ROM. Specifically, we introduce VQ4ALL, a VQ-based technique that leverages codewords to construct various neural networks and facilitate efficient representations. VQ4ALL-constructed low-bit networks achieve state-of-the-art compression quality on popular tasks such as image classification, object detection, and image generation, particularly demonstrating robustness even at extremely high compression ratios. Experimental results show that VQ4ALL achieves compression rates greater than 16×, while maintaining high accuracy across various network architectures, showcasing its advantages in efficient network compression and universal network representation.
{
    \small
    \bibliographystyle{ieeenat_fullname}
    \bibliography{main}
}

\clearpage
\setcounter{page}{1}
\maketitlesupplementary

\section{Ratio Threshold $\alpha$ Searching}

\begin{figure}[tbp]
  \centering
  \includegraphics[width=1\linewidth]{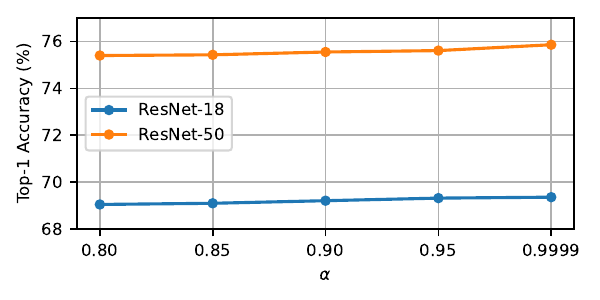} 
    \caption{2-bit ResNet-18/50 compression with varying ratio threshold $\alpha$ values.}
    \label{alpha}
\end{figure}

As shown in Figure \ref{alpha}, we evaluate the impact of different ratio threshold $\alpha$ values in the Progressive Network Construction Strategy on 2-bit ResNet-18/50. The results demonstrate that smaller $\alpha$ values lead to lower network accuracy, indicating that prematurely determining optimal assignments is detrimental to the construction of low-bit networks. Moreover, ResNet-50 is more sensitive to variations in $\alpha$, especially when $\alpha$ decreases from 0.9999 to below 0.95.

\section{Optimal Assignment Distribution}

\begin{figure}[tbp]
  \centering
  \includegraphics[width=1\linewidth]{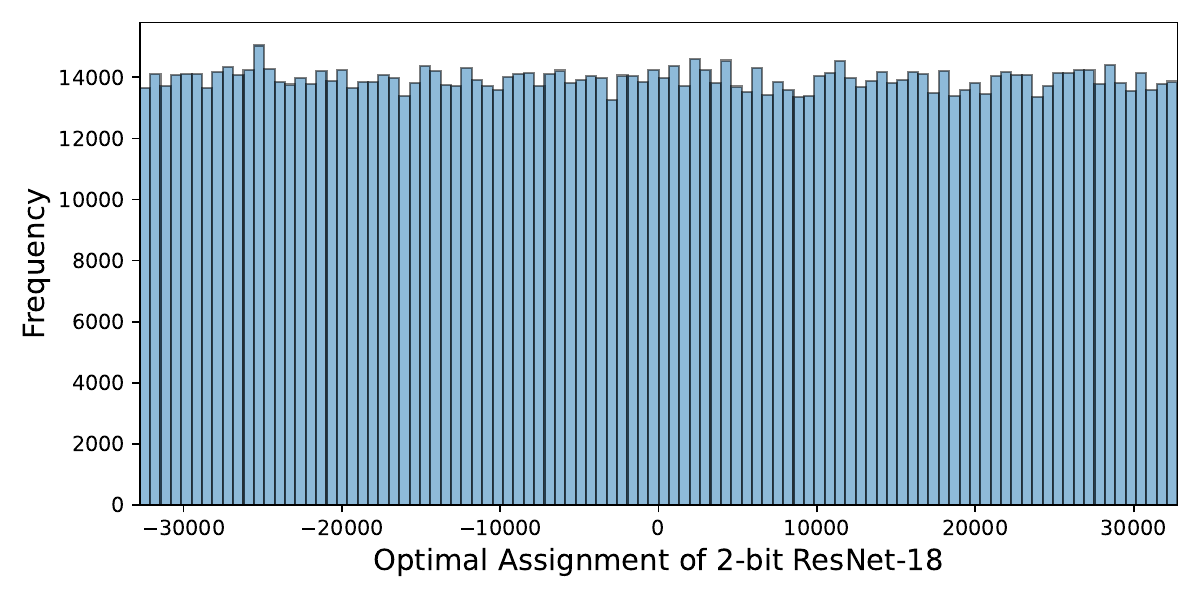} 
  \includegraphics[width=1\linewidth]{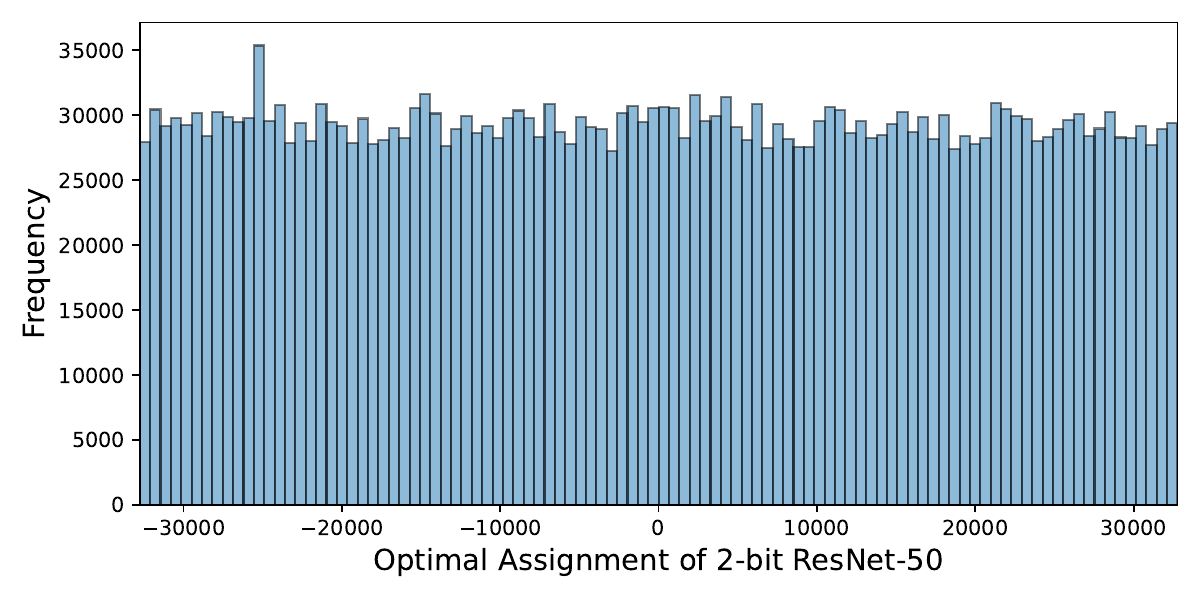} 
  \includegraphics[width=1\linewidth]{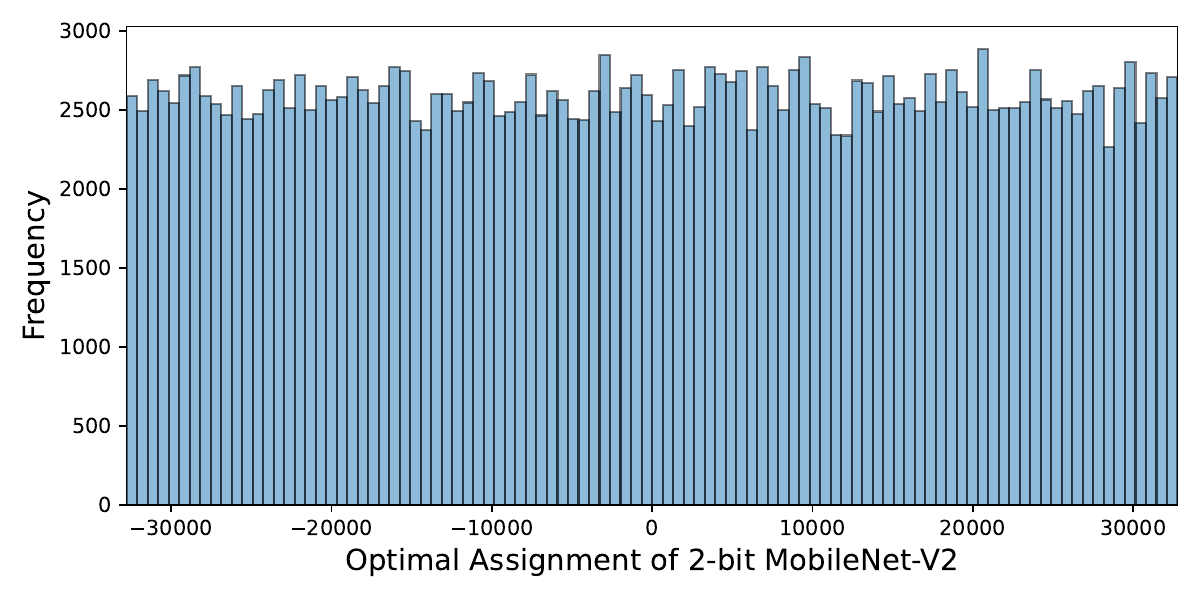} 
  \includegraphics[width=1\linewidth]{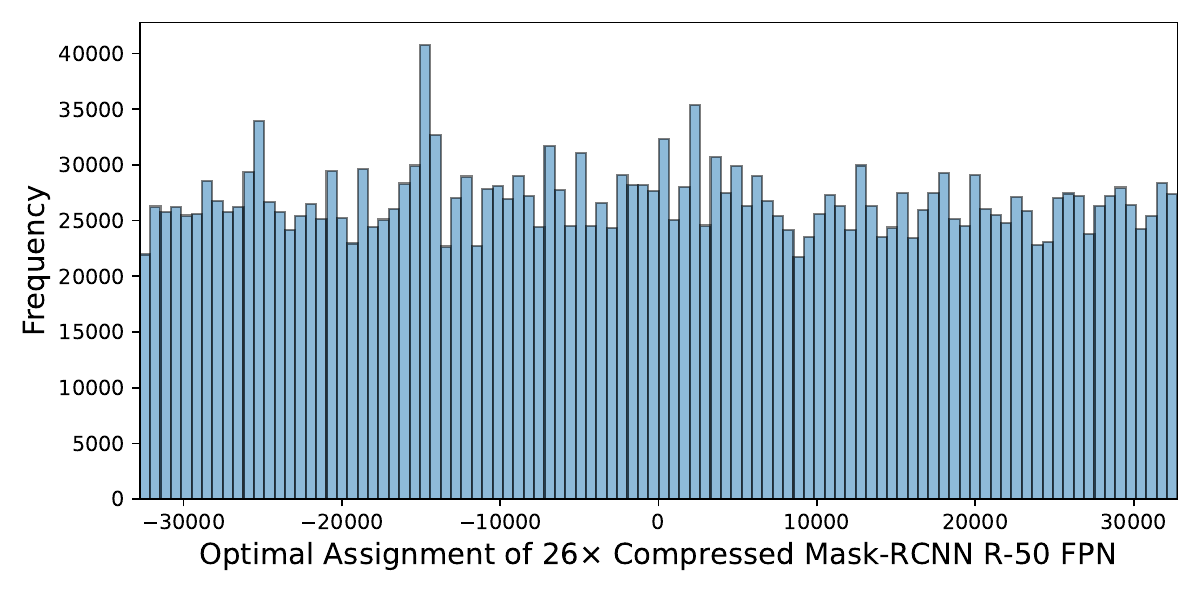} 
  \includegraphics[width=1\linewidth]{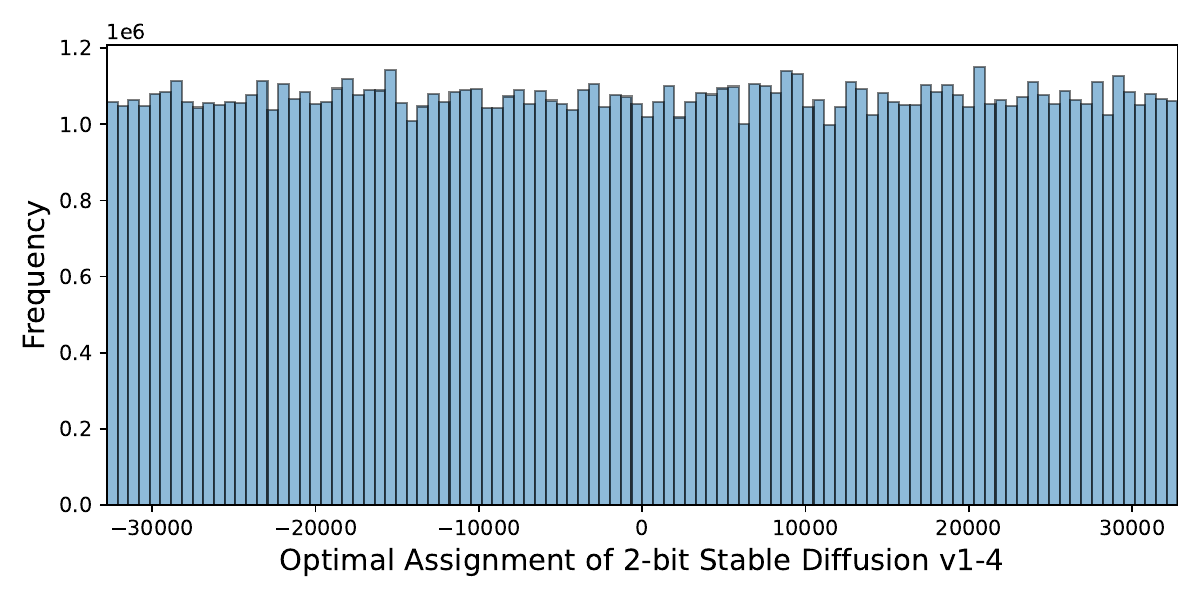} 
    \caption{Optimal assignment distribution of various low-bit Networks, with most layers constructed from the universal codebook.}
   \label{distribution}
\end{figure}

To verify whether the codewords of the universal codebook are fully utilized, we analyzed the optimal assignment distribution across different networks. As shown in Figure \ref{distribution}, each type of low-bit network is evenly composed of different codewords of the same universal codebook. This indicates that the low-bit networks constructed by VQ4ALL do not exhibit a preference for specific codewords, leading to highly efficient information utilization of the universal codebook. We also present the pseudo-code of our proposed VQ4ALL method in Algorithm \ref{algorithm}.

\section{Design of the Universal Codebook}

As shown in Table \ref{universal}, we adopt Kernel Density Estimation (KDE) to fit the combinations of weights from different networks and randomly sample to generate various universal codebooks. We then evaluate the impact of these codebooks on network performance. The results indicate that when the universal codebook is sampled only from the ResNet-18's weights, 2-bit ResNet-18 achieves the best accuracy. As the combination length increases, the network accuracy only slightly decreases. This demonstrates that our VQ4ALL approach does not rely on the similarity between the distribution of the universal codebook and the network's own weights. Instead, it selects optimal codewords from the universal codebook for efficient network representation.

\begin{table}[tbp]
\centering
\begin{tabular}{c|cccccc}
\toprule
   \multicolumn{1}{c}{}   & 1  & 1+2  & 1+2+3 & 1+2+3+4 \\  
   \midrule
    ResNet-18     & 69.41 & 69.39  & 69.37  & 69.36  \\   
    ResNet-50       & 75.83 & 75.92  & 75.88  & 75.86 \\  
    \bottomrule
   
\end{tabular} 
\caption{2-bit ResNet-18/50 compression with various universal codebooks generated from kernel density estimations of different weights combinations. 1: ResNet-18, 2: ResNet-50, 3: Mask R-CNN R-50 FPN, 4: Stable Diffusion v1-4. MobileNet-V2 is excluded from the experiments due to its relatively small parameter size.}
\label{universal}
\end{table}

\section{Configuration of Candidate Assignments}

As shown in Table \ref{init}, we evaluated the impact of different configurations of candidate assignments on network performance. Random candidate assignments yield the poorest performance, with the accuracy of 2-bit ResNet-18 and 2-bit ResNet-50 dropping to only 39.97\% and 44.36\%, respectively. For 2-bit ResNet-50, configuring candidate assignments based on Euclidean distance achieves 0.18\% higher accuracy than cosine similarity, suggesting that VQ4ALL favors maintaining minimal deviation between the quantized feature and the floating-point feature of each layer.

Additionally, we evaluated the initialization strategies for the ratios, comparing the inverse proportional initialization described in Equation 6 with equal initialization. For 2-bit ResNet-18, inverse proportional initialization reduces the number of iterations required for network construction by 15\% and achieves 0.11\% higher accuracy than the equal initialization method.

\begin{table}[tbp]
\centering
\begin{tabular}{c|cccccc}
\toprule
   \multicolumn{1}{c}{}   & Rand  & Cos  & Euclid & Euclid + Init \\  
   \midrule
    ResNet-18     & 39.97 & 69.22  & 69.25  & 69.36  \\   
    ResNet-50       & 44.36 & 75.48  & 75.66  & 75.86 \\  
    \bottomrule
   
\end{tabular} 
\caption{2-bit ResNet-18/50 Compression with Different Candidate Assignment Initialization Methods. 1: Random Initialization. 2: Cosine Similarity. 3: Euclidean Distance. 4: Ratios Initialization for Candidate Assignments Generated based on Euclidean Distance, as Described in Equation 6.}
\label{init}
\end{table}

\section{Settings of Block-Wise Knowledge Distillation}

Section 4.2 introduces the objective function of block-wise knowledge distillation, where different networks are divided into their respective primary blocks based on their structure. For ResNet-18/50, the primary blocks are 'BasicBlock' and 'Bottleneck'. For MobileNet-V2, the primary block is 'InvertedResidual'. For Mask R-CNN R-50 FPN, the primary blocks are 'Bottleneck', 'inner blocks', and 'layer blocks'. For Stable Diffusion v1-4, the primary blocks are 'BasicTransformerBlock' and 'ResnetBlock2D'. The names of all primary blocks are derived from the official code.

\section{Comparison of Generated Images}

\begin{figure*}[tbp]
  \centering
  \includegraphics[width=0.875\linewidth]{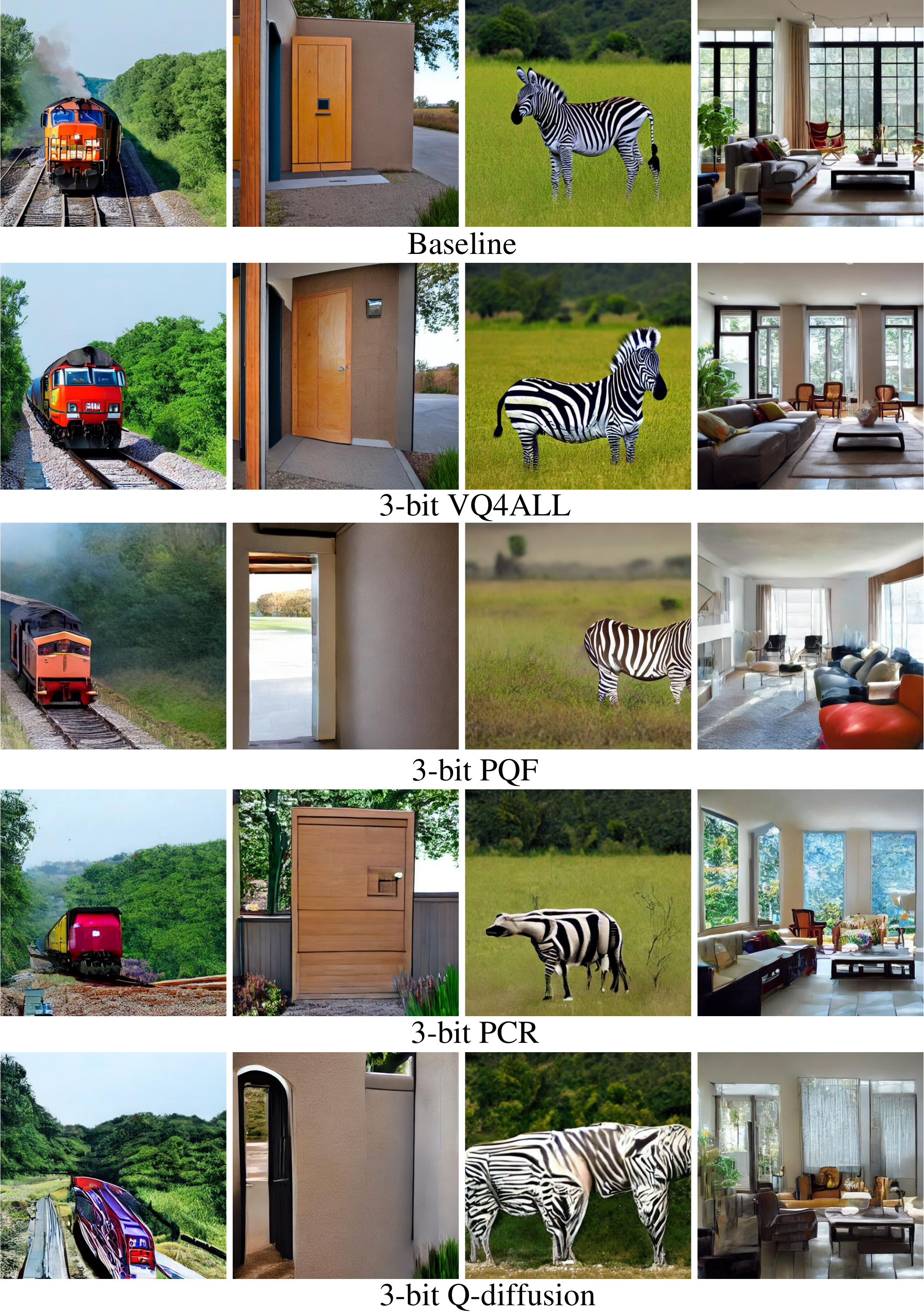} 
    \caption{3-bit Stable Diffusion v1-4 image generation using COCO prompts.}
   \label{3bit}
\end{figure*}

\begin{figure*}[tbp]
  \centering
  \includegraphics[width=0.875\linewidth]{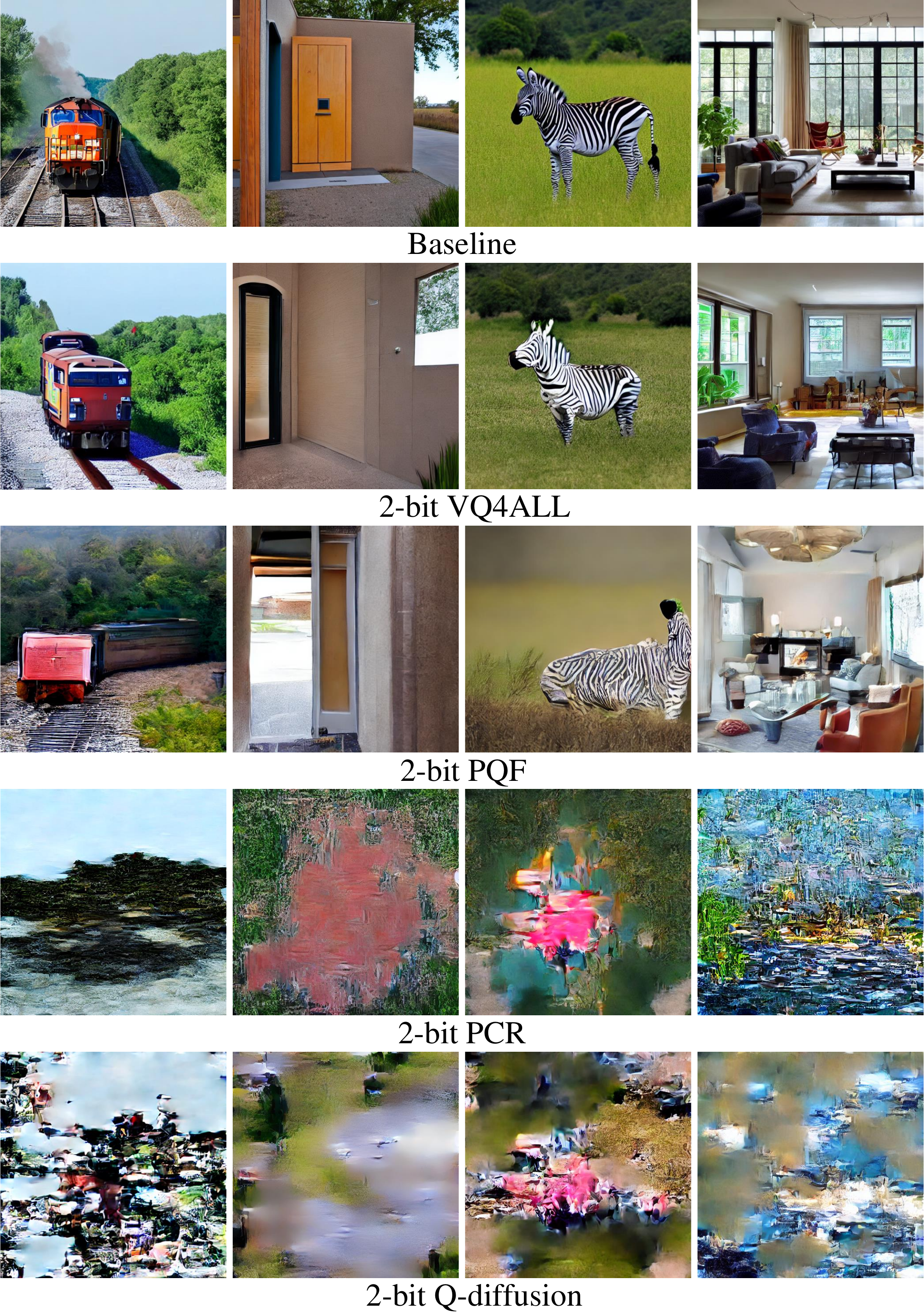} 
    \caption{2-bit Stable Diffusion v1-4 image generation using COCO prompts.}
   \label{2bit}
\end{figure*}

In Figure \ref{3bit} and Figure \ref{2bit}, we present the images generated by the low-bit Stable Diffusion v1-4 network on COCO prompts. Compared to other state-of-the-art uniform quantization methods, the images generated by VQ4ALL are more similar to those produced by the floating-point network. Notably, under 2-bit compression, other methods fail to maintain image quality, whereas VQ4ALL continues to generate realistic and high-quality images.

\renewcommand{\algorithmicrequire}{\textbf{Input:}}
\renewcommand{\algorithmicensure}{\textbf{Output:}}
\begin{algorithm}
\caption{Our proposed VQ4ALL algorithm}
\label{algorithm}
\begin{algorithmic}[1]
\REQUIRE Concatenated weight sub-vectors $W$ of networks
\REQUIRE Codeword number $k$, vector length $d$
\REQUIRE Candidate assignments number $n$
\REQUIRE Calibrate dataset $\{\mathbf{x}, \mathbf{y}\}$
\ENSURE Low-bit networks with universal codebook $C$ and optimal assignments $A$
\STATE \textbf{Initialization of the Universal Codebook:}
\STATE Compute the kernel density estimation (KDE) of $W$
\STATE Randomly sample from the estimation to obtain $C$
\STATE Create $A_c$ and $R$ based on Equation 5 and 6
\STATE \textbf{Progressive Network Construction Strategy:}
\FOR{$\mathbf{x}, \mathbf{y}$}
    \STATE Input $\mathbf{x}$ to original network and low-bit network
    \STATE Generate calibration feature with $W$
    \STATE Generate quantized feature with $\widehat{W} = RC[A_c]$
    \STATE Minimize Equation 12 and update $R$
    \STATE Construct networks by selecting some optimal assignments based on Equation 14
    \IF{All optimal assignments are selected}
        \STATE break
    \ENDIF
\ENDFOR
\end{algorithmic}
\end{algorithm}

\end{document}